\def\year{2020}\relax
\documentclass[letterpaper]{article} % DO NOT CHANGE THIS
\usepackage{aaai20}  % DO NOT CHANGE THIS
\usepackage{times}  % DO NOT CHANGE THIS
\usepackage{helvet} % DO NOT CHANGE THIS
\usepackage{courier}  % DO NOT CHANGE THIS
\usepackage[hyphens]{url}  % DO NOT CHANGE THIS
\usepackage{graphicx} % DO NOT CHANGE THIS
\usepackage{graphicx}  % DO NOT CHANGE THIS
\usepackage{times}
\usepackage{soul}
\usepackage[utf8]{inputenc}
\usepackage{graphicx}
\usepackage{amsmath}
\usepackage{amsthm}
\usepackage{booktabs}
\usepackage{multirow}
\usepackage{dsfont}
\usepackage{paralist}
\usepackage[linesnumbered, ruled, vlined]{algorithm2e}
\usepackage{algpseudocode}
\usepackage{color}
\usepackage{dsfont}
\usepackage{url}

\title{Parallel Computation of Graph Embeddings}
\author{Chi Thang Duong\textsuperscript{\rm 1} Hongzhi Yin\textsuperscript{\rm 2} Thanh Dat Hoang\textsuperscript{\rm 3} Truong Giang Le Ba\textsuperscript{\rm 3}\\ \Large \textbf{Matthias Weidlich}\textsuperscript{\rm 4} \textbf{Quoc Viet Hung Nguyen}\textsuperscript{\rm 5} \textbf{Karl Aberer}\textsuperscript{\rm 1}\\ \Large %\textbf{AAAI Style Contributions by
\textsuperscript{\rm 1} EPFL %If you have multiple authors and multiple affiliations  use superscripts in text and roman font to identify them. For example, Sunil Issar,\textsuperscript{\rm 2} J. Scott Penberthy\textsuperscript{\rm 3} George Ferguson,\textsuperscript{\rm 4} Hans Guesgen\textsuperscript{\rm 5}. Note that the comma should be placed BEFORE the superscript for optimum readability
\textsuperscript{\rm 2} The University of Queensland\\
\Large
\textsuperscript{\rm 3} HUST
\textsuperscript{\rm 5} Griffith University
\textsuperscript{\rm 4} Humboldt-Universität zu Berlin
}

\newcommand{\sstitle}[1]{\smallskip\noindent\textbf{#1.\/}}

\newtheorem{mythm}{Theorem}
\newtheorem{mylmm}{Lemma}
\newtheorem{myprob}{Problem}

\begin{document}
%\title{Efficient Graph Embedding under Resource Constraints}

\maketitle
\begin{abstract}
Graph embedding aims at learning a vector-based representation of
vertices that incorporates the structure of the graph. This
representation then enables inference of graph properties.
Existing graph embedding techniques, however, do not scale well to large
graphs. We therefore propose a framework for parallel computation
of a graph embedding using a cluster of compute nodes with resource
constraints. We show how to distribute any existing embedding technique by
first splitting a graph for any given set of 	
constrained compute nodes and then reconciling the embedding spaces derived
for these subgraphs.
%that is able to leverage multiple machines with different resource constraints
%to construct graph embedding. Our framework first partitions the graph into
%several overlapping subgraphs that can be handled on different machines
%according to their resources. Then, the embeddings for the subgraphs are
%constructed and reconciled based on their common nodes to obtain a final
%embedding for the whole graph.
We also propose a new way to evaluate the quality of graph embeddings that is
independent of a specific inference task. Based thereon, we give a formal bound on the difference between
the embeddings derived by centralised and parallel computation.
Experimental results illustrate that our approach for parallel computation scales well, while largely maintaining the embedding quality.
%achieving that our framework achieves comparable
%performance with unconstrained approach while respecting the resource
%constraints.
%The code and datasets are public available at~\cite{}.
\end{abstract}


\section{Introduction}
Graphs are a natural representation of relations between entities in
complex systems, such as social networks or information networks. To enable
inference on graphs, a \emph{graph
embedding} may be learned. It comprises \emph{vertex embeddings},
each being a vector-based representation of a graph's vertex that
incorporates its relations with other
vertices~\cite{hamilton2017representation}.
%one needs to embed each node by some
%hand-crafted features based on some expert knowledge, which makes this a
%tedious and error-prone process.
%Graph embedding, which is a technique to construct distributed representation
%of nodes, allows to automate this process.
%Each node in the graph is
%represented by a vector (i.e. \emph{an embedding}) that captures its
%relationship with other nodes in the graph.
Inference tasks, such as vertex classification and link prediction, can then be
based on the graph embedding rather than the original graph.
Various techniques to learn a graph embedding have been
proposed~\cite{perozzi2014deepwalk,hamilton2017representation,hamilton2017inductive}.
However, common techniques aim at high embedding quality at the expense of computational efficiency, so that they do not scale to extremely large graphs that comprise billions of nodes and trillions of edges~\cite{lerer2019pytorch}. Computing an embedding with traditional techniques for graphs of such sizes may take weeks, which renders it practically infeasible. Moreover, such computations are challenging in terms of their memory requirements. For example, storing a graph with two billion nodes in main memory requires around 1TB RAM~\cite{lerer2019pytorch}, which exceeds the capacity of commodity servers.

%Large amounts of memory are needed to store a graph for processing and the actual learning of vertex embeddings may take days or even weeks for real-world graphs~\cite{hamilton2017inductive}.

%\edit{Modern graphs can be extremely large, with billions of nodes and trillions of edges. Standard graph embedding methods don’t scale well out of the box to operate on very large graphs. There are two challenges for embedding graphs of this size. First, an embedding system must be fast enough to allow for practical research and production uses. With existing methods, for example, training a graph with a trillion edges could take weeks or even years. Memory is a second significant challenge. For example, embedding two billion nodes with 128 float parameters per node would require 1 terabyte of parameters. That exceeds the memory capacity of commodity servers.}\cite{lerer2019pytorch}

%while assuming unlimited resources required to construct the
%embeddings. This limits the scalability and applicability of graph embedding
%techniques. For instance, several techniques require a large amount of memory
%to store and process the graph. In addition, the larger the graph is, the
%longer the training time required to obtain a good graph embedding. Several
%graph embedding techniques require days or even weeks to
%train\cite{hamilton2017inductive}.

Against this background, we argue that graph embedding
shall exploit parallel computation in a cluster of compute nodes. However,
the parallelisation of graph embedding is challenging for various
reasons.

First, a graph shall be split into subgraphs, each of which is processed
separately by a compute node. This is difficult, since compute nodes have
resource constraints. Their memory capacity is limited, which induces a bound
on the size of any subgraph that can be handled by a node. Any split of the
graph thus needs to incorporate these resource constraints.

A second challenge is the reconciliation of embeddings computed for subgraphs.
Each node learns an embedding for only a subset of graph vertices, so that the results are vectors in different embedding spaces. An
alignment of these spaces is needed to obtain a meaningful
graph embedding.

The need to reconcile embedding spaces leads to a third challenge, i.e., the evaluation of the quality of a graph embedding. Traditionally, embedding quality is assessed based on some inference task, such as vertex
classification or link prediction. However, such an approach renders the measurement dependent on a specific task, which precludes a general assessment of how the quality of embeddings is affected by parallel computation and subsequent reconciliation.

%, while at the same time striving for maximal utilization of compute nodes
%without comprising the

%we consider the problem of designing a graph embedding framework that also
%takes into account the resource constraints such as time and memory. Relaxing
%the unlimited resource assumption, we assume access to a pool of machines
%where
%each has a limited amount of memory. Given this constraint, we want to
%maximize
%the utilization of all the machines while reducing the difference in quality
%between embeddings created with unlimited and limited resource.

In this paper, we answer these challenges in a unifying framework. We
provide a method to decompose a graph, such that the resulting subgraphs
satisfy resource constraints imposed by compute nodes. These subgraphs share
some vertices, called \emph{anchors}, that facilitate
reconciliation of the derived embeddings. Intuitively, the differences in the
embeddings of an anchor as part of different subgraphs guide the alignment of
the respective embedding spaces. We show how to select such anchors effectively
in order to minimise the impact of the parallel computation on the
embedding quality.
To assess this effect, we adopt the idea of the Pairwise Inner Product
(PIP)~\cite{yin2018dimensionality,yin2018global}, a loss metric proposed for
word embedding tasks, and propose a way to measure embedding quality in an intrinsic manner. 
Based thereon, we give a theoretical bound on the
difference between a graph embedding derived by centralised
computation and our parallelization scheme. Comprehensive
experiments with real-world data illustrate the effectiveness and efficiency of our approach.

In the remainder, Section~\ref{sec:model} first gives an overview of our approach.
Section~\ref{sec:anchor} then outlines how to embed subgraphs, while
Section~\ref{sec:recon} focuses on reconciliation of embedding spaces. Section~\ref{sec:pip} presents a measure to compare embeddings, which is used for a theoretical analysis in
Section~\ref{sec:theory}. Experimental
results are presented in Section~\ref{sec:exp}. Section~\ref{sec:related}
reviews related work, before we conclude in Section~\ref{sec:con}.

\section{Parallel Graph Embedding Framework\label{sec:model}}

%\subsection{Graph Embeddings and Computational Model}
\subsection{Computational Model}
%\sstitle{Graph embedding}
Let $G = (V,E)$ be an undirected graph with vertices $V$ and edges
$E\subseteq [V]^2$.
A graph embedding technique aims to learn a mapping
$f: V \rightarrow \mathcal{R}^d$ from vertices $V$ to an embedding in a
low-dimensional space ($d \ll |V|$), such that `similar' vertices are mapped to
close vertex embeddings~\cite{hamilton2017representation}.
%A decoder is a function that takes embeddings as inputs. Its output is application-dependent and it allows us to evaluate the quality of the embeddings.
%For instance, an output of a decoder is the similarity between two embeddings.
%Traditionally, we expect the embeddings to capture the structural relationship between the nodes. More precisely, we expect the homophily principle is preserved where embeddings of nodes that are close in the network to also be close in the embedding space.
The similarity of vertices may, e.g., be measured based on the ratio of
common neighbouring vertices or the similarity of vertex labels.
%There are several ways of defining similarity between nodes in a graph. For
%instance, neighbor nodes are considered to be similar or nodes that have the
%same set of neighbors or node labels are similar.
Closeness of two vertex embeddings, in turn, is commonly measured by their
cosine similarity. Typically, the mapping $f$ is parameterised, so
that the graph embedding problem is to find parameters of $f$ that minimise the
difference between the similarity assigned to vertices and the closeness of
their respective embeddings.
%As such, we assess the quality of an embedding by the degree to which the
%output of a decoder meets our expectations, which is formalised by a
%\emph{loss
%function} $\mathit{L}$. As encoders and decoders are usually parametrised
%functions, finding a good network embedding becomes the problem of setting the
%encoder and decoder parameters, such that the loss function is minimised.
%For instance, in community detection, the output of the decoder are community labels for the nodes and we expect the obtained labels to reflect real-world communities.
%the correctness of the detected community is a measure of the embedding quality.

%\sstitle{Resource constraints}

Traditional techniques focus on optimising the quality of a graph embedding, which limits their scalability.  Learning embeddings for large graphs comes with a large memory footprint, while training times may
extend to days or even weeks~\cite{hamilton2017inductive}. We therefore aim to
parallelise the computation of a graph embedding in a cluster of $n$ compute
nodes. In a cluster, each such node has resource constraints in terms of the
available memory and runtime. Here, we focus on memory constraints, since
runtime requirements directly depend on the size of the graph and, hence, the
required memory to store the graph. Specifically, we denote by $k_1,\ldots,k_n$
the memory constraints of $n$ compute nodes and assume that they are expressed
as the maximal size of the graph that can be handled, i.e., the $i$-th node may process a graph with up to $k_i$ vertices.

\subsection{Problem Statement}

To parallelise the computation of a graph embedding, the following problem needs
to be addressed:

\begin{myprob}
\label{prob:para}
%Given a graph $G = \{V,E\}$, a reference embedding procedure $\mathbf{Z}_n = f(G)$ and a node constraint $k$, the problem we want to solve is to devise an embedding procedure $f_p$ such that
Let $G = (V,E)$ be a graph and $f$ a reference embedding technique, so that
$f(G)= \mathbf{E}$. Given a cluster of $n$ nodes with memory constraints
$k_1,\ldots,k_n$, the problem of parallel graph embedding is to devise an
embedding technique $f_d$, $f_d(G)=\mathbf{F}$, that exploits the $n$ nodes,
satisfies the memory constraints of all nodes, and minimises the difference
between $\mathbf{F}$ and $\mathbf{E}$.
%$\mathbf{Z}_n$ is the node embeddings obtained using a non-parallel embedding procedure $\mathbf{Z}_n = f(G)$ and $\mathbf{Z}_p$ is the node embeddings obtained using a parallel embedding procedure $\mathbf{Z}_p = f_p(G)$, the problem we want to solve is to find $f_p$ such that the difference between $\mathbf{Z}_p$ and $\mathbf{Z}_n$ to be minimal.
%a non-parallel embedding procedure $f: G \rightarrow \mathds{R}^d$, a parallel embedding procedure $f_p$, the problem we want to solve is to find $f_p$ such that the quality of.
\end{myprob}

\noindent
Here, we assume that $\sum_{1\leq i\leq n} k_i > |V|$.
%Also, in the remainder, we consider homogeneous clusters ($k_1 = \ldots = k_n
%=
%k$) and heterogeneous clusters (memory constraints differ per node).
%In addition, we
%assume that the machines are similar, which means their constraints are the
%same: $k_1 = ... = k_n = k$. We relax this assumption in the following
%section.
As a reference embedding technique, we consider techniques for
centralised computation of graph embeddings, such as
DeepWalk~\cite{perozzi2014deepwalk}, HOPE~\cite{ou2016asymmetric},  and SGC~\cite{wu2019simplifying}. These techniques are representative methods for the three main approaches to graph embedding, which are grounded in random-walks, matrix factorisation, and graph neural networks, respectively. 
%... The node constraint $k_i$ encompasses not
%only the memory but also time constraint as the memory requirement of an
%embedding procedure $f$ increases with the graph size while its training time
%is also longer when the graph is larger\cite{hamilton2017inductive}. %\todo{In
%Section~\ref{}, we perform a preliminary analysis to show the correlation
%between the graph size and the memory requirement and training time. This
%shows
%the need for leveraging multiple machines to handle massive graphs. }

%In the above problem, $\mathbf{Z}_n$ is called the reference embedding while $\mathbf{Z}_p$ is the parallel embedding. The non-parallel embedding procedure $f$ can be any of existing graph embedding technique such as DeepWalk\cite{}, node2vec\cite{}... In the following, we discuss our proposed parallel graph embedding framework.

\subsection{Approach}

Our general idea is to split a graph into $n$ subgraphs, so that the
computation of the vertex embeddings can be parallelised among $n$ compute
nodes. Then, the memory constraint $k_i$ of the $i$-th node gives a bound for
the size of the $i$-th subgraph. However, the vertex embeddings computed per
subgraph are of different
spaces. Our approach therefore is to construct subgraphs that share some
vertices,
called \emph{anchors}, which enable us to reconcile the embedding
spaces.

A realisation of this idea is challenging, though. There is an exponential
number of possible decompositions of a graph into $n$ partially overlapping
subgraphs. Each split has different consequences for the quality of
the reconciled embedding. To minimise the difference between the graph
embeddings obtained by centralised and parallel computation,
an effective strategy for the selection of anchors is needed.

\begin{figure*}[h!]
	\centering
	\includegraphics[width=.85\linewidth]{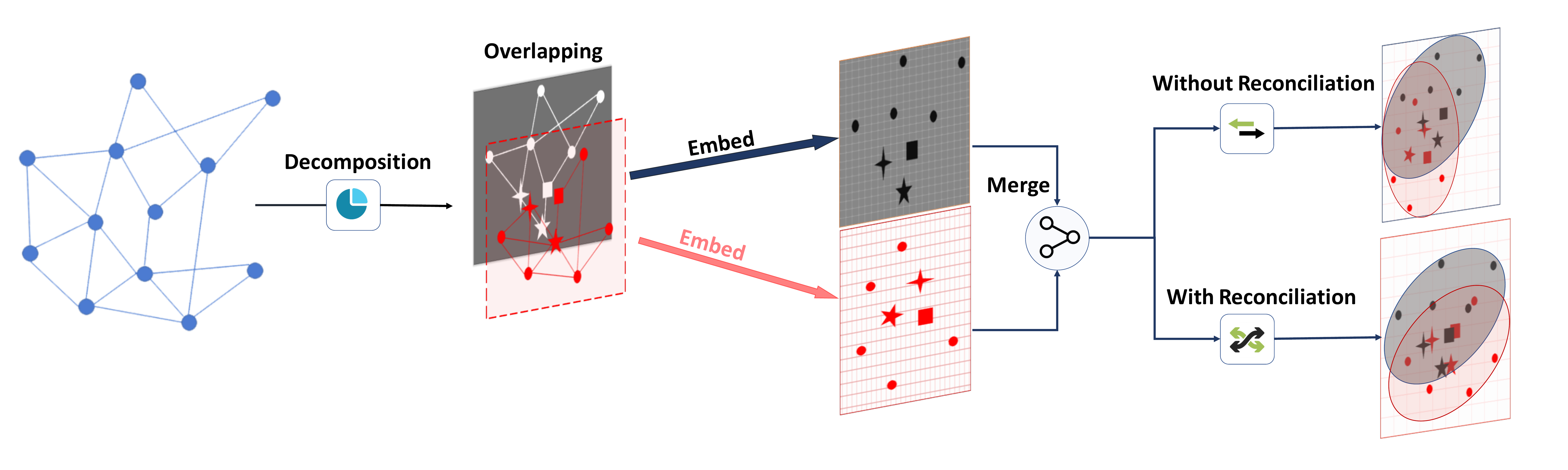}
	\vspace{-1em}
	\caption{Framework for parallel computation of graph embeddings on distributed nodes with resource constraints.}
	\label{fig:archi}
	\vspace{-1em}
\end{figure*}

%Our parallel graph embedding framework involves 3 components as follows. Figure~\ref{} illustrates this process.
Given the above considerations, we propose a graph embedding framework, shown
in Figure~\ref{fig:archi}, as follows:

\begin{compactitem}
\item[\emph{Select anchors.}] First, we decompose
the graph into subgraphs and select a set of anchors. Given the importance
of anchors for reconciliation of embedding spaces, we present an effective
selection strategy in Section~\ref{sec:anchor}.
\item[\emph{Embed subgraphs.}] Once a graph is decomposed into $n$
subgraphs, we learn an embedding per subgraph on
a separate compute node.
%As a result, given a graph embedding technique $f$, we obtain an embedding for
%each subgraph.
As our framework is agnostic to the chosen embedding algorithm, it improves
scalability of \emph{any} existing graph embedding technique.
%\todo{It is worth noting that the selected embedding technique should be
%similar to the non-parallel graph embedding technique to make $\mathbf{Z}_n$
%and $\mathbf{Z}_p$ comparable.}
\item[\emph{Reconcile embedding spaces.}] %As the embeddings for the
%subgraphs are obtained independently, the embeddings
%may belong to different embedding spaces.
In this step, we reconcile the embedding spaces obtained for different
subgraphs on different compute nodes by constructing a function that maps from
one embedding space to another one.
%Inspired by work on network
%alignment~\cite{man2016predict} and
%cross-language dictionary building~\cite{conneau2017word,artetxe2016learning},
We show how to construct such a mapping in Section~\ref{sec:recon}.
\end{compactitem}

% Given the above proposed framework, Problem~\ref{prob:para} can be rewritten as follows:

% \begin{myprob}
% Given a graph $G = \{V,E\}$, a non-partition embedding $\mathbf{Z}_n$, an embedding function $h: G \rightarrow \mathds{R}^d$, a mapping procedure $m$ and a predefined number $n$, the problem we want to solve is find a set of $n$ anchors nodes $A \subset V$ such that the difference between $\mathbf{Z}_n$ and the embedding obtained by the parallel embedding process $\mathbf{Z}_p = f_p(G,h,m,n)$ is minimized.
% \end{myprob}
%In Problem~\ref{prob:para}, we aim to minimize the difference between
%$\mathbf{E}$ and $\mathbf{F}$. As a result, we need a way to quantify their
%difference. In the following, we give an overview of how to evaluate the
%embedding before discussing it in detail in Section~\ref{sec:pip}.

%\sstitle{Evaluating embedding quality}
\noindent
As stated in Problem~\ref{prob:para}, we
strive for a minimal distance between the graph embedding obtained with
centralised computation,~$\mathbf{E}$, and the one
derived by parallel computation,~$\mathbf{F}$.
We therefore also clarify how to evaluate embeddings independent of a specific
inference task (Section~\ref{sec:pip}) 
and give a bound for the difference
between centralised and parallel
computation of graph embeddings (Section~\ref{sec:theory}).

\section{Decomposition for Subgraph Embedding}
\label{sec:anchor}

We first outline requirements on the graph decomposition for 
parallel embedding. We then present a strategy for anchor selection, 
before discussing the embedding of subgraphs.
%   component. There are several ways to select anchor nodes. Intuitively, we 
%want to select ``important'' nodes to be anchors. The importance of a node can 
%be considered w.r.t the partitions or not. As a result, we can either 
%partition 
%and then select the anchors or vice versa. In the following, we first discuss 
%the graph partitioning component and then the anchor selection strategy. 

\subsection{Graph Decomposition}
While graph decomposition is a well-studied problem, there are several 
requirements that pertain to our setting. First, subgraph sizes shall be 
proportional to the memory constraints of compute nodes. 
%This yields an 
%unbalanced decomposition for a heterogeneous cluster, while subgraph sizes 
%shall be equal for a homogeneous cluster.
%as the machines are uniform, 
%in order to leverage the machines to their fullest, the partition sizes should 
%be equal. The uniform assumption can be relaxed by considering a graph 
%partitioning technique that can partition graphs into unequal sizes. 
%Second, if we perform partitioning after selecting the anchor nodes, the partitioning is only done on the induced graph after removing the anchor nodes. More precisely, let $A$ be the set of anchor nodes, then we only partition the graph $G'$ that is the induced graph from the set of nodes $V\setminus A$. 
Second, the number of edges between non-anchor vertices shall be minimised, as 
these edges are neglected when embedding the subgraphs. 
Formally, a graph decomposition is a function that takes a graph $G$, the 
number of subgraphs $n$, and the maximal subgraph sizes $\{k_1,\ldots,k_n\}$ as 
input, and returns a set of subgraphs, $\mathit{decompose}(G, n, 
\{k_1,\ldots,k_n\}) = 
\{P_1, \ldots,P_n\}$, where $P_i = (V_i, E_i)$ is a subgraph of $G$. 
%In case of 
%a balanced decomposition with equal subgraph sizes, we write the decomposition 
%as $d(G, n, k)$.
%, the partition 
%size is a set of constraints: $P_1, P_2,...,P_n = p(G', n, 
%\{k'_1,k'_2,...,k'_n\})$. %It is worth noting that if we select the anchors 
%before partitioning, the graph $G'$ is an induced graph after removing the 
%anchor nodes $A$ while the constraint needs to consider the anchor nodes as 
%well: $k' = k - |A|$ or $k'_i = k_i - |A|$.
Here, we leverage METIS~\cite{karypis1998software} to 
decompose a graph. It is a scalable algorithm for unbalanced decomposition, 
i.e., it incorporates constraints on maximal subgraph 
sizes.

%While anchor node selection for graph embedding is a new problem, graph partitioning is a well-researched problem. 

\subsection{Anchor Selection}
%Given the above reconciliation component and a fixed graph embedding technique, the only component that can affect the performance of our paralell embedding process is the anchor selection component. As a result, given a fixed reconciliation component and a fixed embedding technique, Problem~\ref{prob:para} can be rewritten as follows:
%\todo{As the number of nodes in $G$ is very high, the problem is computational expensive. As a result, we propose several heuristics to solve the above problem. }

Several criteria shall be considered when devising a strategy to select a set 
of anchors $A\subseteq V$ for the subgraphs of a graph $G=(V,E)$. The 
number of anchors $|A|$ has a great effect on the ability to reconcile the 
embeddings computed for different subgraphs. A large number of 
anchors provides many clues for reconciliation and, thus, is expected to lead 
to higher overall embedding quality. 
%In the extreme case, if all 
%vertices are anchors, the mapping between embedding spaces is perfect as the 
%spaces are defined by the same dimensions. 
%As discussed above, as the embedding space is in high dimension and there are a large number of nodes, we need a large amount of anchor nodes to construct good mapping.
On the other hand, a large number of anchors is problematic, as the respective 
vertex embeddings are computed several times. As such,
there is a trade-off between the ability to reconcile embedding spaces (i.e., 
the goodness of the mapping between the spaces) and the overall required 
learning effort.
%Second, to maximize the efficiency when training in parallel, the subgraphs should have nearly the same number of nodes.
Moreover, the structural importance of anchors 
influences the reconciliation of embedding spaces. A higher 
anchor connectivity yields a better alignment of the 
dimensions of embedding spaces. 
Against this background, we propose to select anchors by exploiting properties of the decomposition of the graph. Given the set of 
vertices that are part of subgraph borders, denoted by $U\subseteq V$, 
we select a set of anchors $A \subseteq U$ based on vertex connectivity. That 
is, we select the top-$d$ nodes in $U$ that have 
the highest number of edges between subgraphs. 
The intuition is that by 
selecting these vertices, we are able to 
reduce the number of edges that are removed by the decomposition and anchor 
selection process. 

%\vspace*{-.3cm}
\begin{algorithm}
	\footnotesize
	\caption{Graph decomposition for embedding.}
	\label{alg:partition2}
	\SetKwInOut{Input}{input}%
	\SetKwInOut{Output}{output}%
	\Input{Graph $G=(V,E)$; number of subgraphs $n$; number of anchors $d$;\newline 
	maximal subgraph sizes $\{k_1,\ldots,k_n\}$.
	}
	\Output{
		$n$ subgraphs $P_1, \ldots, P_n$; anchors $A$.
	}
	\BlankLine
	\tcp{Graph decomposition}
	$(V_1, E_1),...,  (V_n, E_n) = \mathit{decomp}(G, n, 
	\{k_1,...,k_n\})$\;
	    \label{alg:partition2:part}	
	\BlankLine
	\tcp{Select anchors}
	$C = \{ \{u,v\} \in E \mid u \in V_i  \wedge v \in V_j \wedge i \neq 
	j\}$\;
		\label{alg:partition2:init}	
	$L = \{ \{u, h_u\} \mid  h_u = \sum_{v \in V} \mathds{1}_{\{u,v\} \in 
	C}\}$\;
	    \label{alg:partition2:count}	
	$A = select\_top(L, d)$\;
	    \label{alg:partition2:anchors}
	\BlankLine
	\tcp{Merge with anchors}
	\For{$i\in [1,n]$}{
		\label{alg:partition2:merge}
		\vspace{-2em}	
	    \begin{multline*}
	    P_i = (V_i \cup A, \{ \{u,v\} \mid \{u,v\} \in E \ \wedge\\ 
		u\in V_i \cup A \ \wedge \ v\in V_i\cup C\})
	    \end{multline*}
		\vspace{-2em}	
	    %$P_i = (V_i \cup A, \{ \{u,v\} \mid \{u,v\} \in E \ \wedge$\newline  
	    %$u\in V_i \cup A \ \wedge \ v\in V_i\cup C\})$\;
	}
	\Return $P_1, \ldots, P_n, A$\;
\end{algorithm}
%\vspace*{-.3cm}

This strategy is formalised in Algorithm~\ref{alg:partition2}. The algorithm first decomposes the graph $G$ into $n$ subgraphs that satisfy the size constraints 
$\{k_1,\ldots,k_n\}$ (line~\ref{alg:partition2:part}). 
We then consider the border vertices of the resulting subgraphs (line~\ref{alg:partition2:count}) and 
select the $d$ vertices with the highest numbers of 
edges (line~\ref{alg:partition2:anchors}) as anchors. The anchors are added to 
the subgraphs (line~\ref{alg:partition2:merge}), before returning the subgraphs and anchors. 
%\todo{Check xem co add anchor nodes vao cac partition}

%\sstitle{Choosing number of anchors} \todo{Bayesian optimization.}

\subsection{Subgraph Embedding} 
Given the subgraphs $P_1, \ldots ,P_n$ generated above, we leverage any 
existing graph embedding technique for the 
subgraphs. A reference embedding procedure $f$ embeds each 
subgraph $P_i$, which yields a matrix $\mathbf{F}_i$ of vertex embeddings. 
The resulting embedding spaces represented by these matrices tend to differ for 
any two subgraphs. However, since the subgraphs share anchors, it is possible 
to reconcile the embedding spaces. 
\section{Reconciling Embedding Spaces}
\label{sec:recon}

To reconcile embeddings that
have been constructed independently of each other, one may 
either construct a pairwise mapping between any two spaces, or map all
of them into one pivot space. The former has to cope with a combinatorial 
blow-up of the number of required mappings. 
Reconciliation with a pivot space, in turn, requires the joint construction of 
$n$ mappings, with $n$ being the number of subgraphs.

Regardless of the above design choice, our idea is to reconcile embedding 
spaces based on anchors. An anchor will be associated with different embeddings 
in different embedding spaces, even though it relates to the same entity, 
i.e., the same vertex in the original graph. Hence, anchors tell us how to 
convert an embedding space into another one.
%Informally, based on these ``bridges'', we want to ``rotate'' one embedding
%space to another embedding space such that the anchor nodes are aligned.

\begin{figure}[h!]
%	\vspace{-.7em}
	\centering
	\includegraphics[width=.9\linewidth]{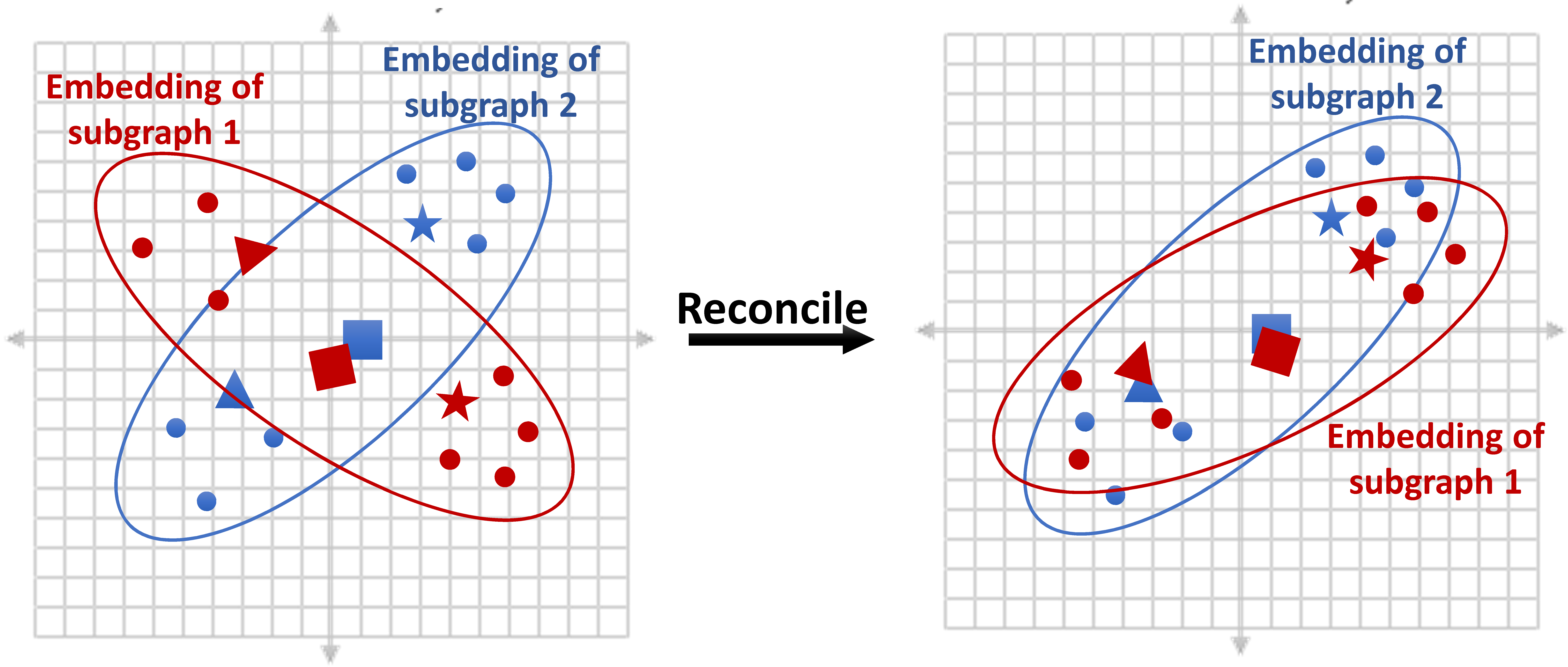}
	\vspace{-.7em}
	\caption{Reconciliation of embedding spaces.}
	\label{fig:parallel}
	\vspace{-.7em}
\end{figure}

%\begin{example}
Figure~\ref{fig:parallel} shows embedding spaces derived for two 
subgraphs, where embeddings of three anchors are shown as stars, 
triangles, and squares. By `rotating' one embedding space, such 
that the embeddings of anchors in either spaces are close to each other, we are 
able to align the two spaces.
%As a side effect, non-anchor nodes are also aligned based on the alignment on
%anchor nodes.
%\end{example}

%In the above example, we are able to align the two embedding spaces using a
%small number of anchor nodes.  It is worth noting that as the embedding space
%are in high dimension, we need to have a large number of anchor nodes to
%correctly align two embedding spaces. In the following, we discuss our
%approach
%to construct such ``rotation''.

%Our parallel embedding computation involves two steps. In the first step, the large graph is partitioned into several smaller subgraphs. In the second step, an embedding mapping technique is used to mend the embedding spaces computed from the subgraphs. In the following, we discuss these steps in detail.

%\sstitle{Learning embedding mapping}
Assume that we decomposed a graph $G$ into subgraphs $P_1,\ldots, P_n$ and 
obtained the embeddings $\mathbf{F}_1,\ldots,\mathbf{F}_n$ for these subgraphs.
%---we later discuss how to find such a partitioning. 
We strive for a mapping of embeddings $\mathbf{F}_1, \ldots, \mathbf{F}_n$ into 
a single embedding space, which, without loss of generality, is one of the $n$ 
spaces, denoted by $\mathbf{F}_0$. 
We propose to learn a mapping function 
$h(\mathbf{F}_i)$ that takes a
source embedding space as input and returns a mapped space, such that
the embeddings of the anchors are close in $\mathbf{F}_0$. This approach is
inspired by techniques for network alignment~\cite{man2016predict} and 
cross-lingual dictionary building~\cite{conneau2017word,artetxe2016learning}. 
The mapping function can be linear or a multilayer
perceptron~\cite{ruck1990multilayer}. In any case, our objective is captured by the following loss function:
\begin{equation}
\label{equ:map}
L(h,\mathbf{F}_i, \mathbf{F}_0, A) = \sum_{v \in A} ||h(\mathbf{z}_{i,v})  - \mathbf{z}_{0,v}||_F
\end{equation}
where $||.||_F$ is the Frobenius norm, $A$ is the set of anchors, and
$\mathbf{z}_{i,v}$ is the embedding of anchor $v$ in $\mathbf{F}_i$.
Since it was shown that a linear function is sufficient to obtain a good 
mapping~\cite{conneau2017word,artetxe2016learning}, we define the mapping 
function as $h(\mathbf{F}_i) = \mathbf{F}_i \times \mathbf{W}$ where 
$\mathbf{W} \in \mathcal{R}^{d\times d}$ and $d$ is the embedding dimensionality. 
The above equation is rewritten in its matrix form, if we denote the embedding 
matrices of the anchor nodes of $\mathbf{F}_i$ and $\mathbf{F}_0$ as 
$\mathbf{H}_i$ and $\mathbf{H}_0$:
\[L(\mathbf{H}_i, \mathbf{H}_0, \mathbf{W}) = ||\mathbf{H}_i\mathbf{W} - \mathbf{H}_0||_F\]
Furthermore, it is known that a better mapping is obtained when enforcing 
orthogonality on $\mathbf{W}$~\cite{conneau2017word}. Under the orthogonality 
constraint, the mapping matrix $\mathbf{W}$ that minimises Eq.~\ref{equ:map} is 
found using singular value decomposition (SVD). Let 
$\mathbf{U}\mathbf{\Sigma}\mathbf{V}^T = \mathbf{H}_0\mathbf{H}_i^T$ be the 
SVD of the matrix $\mathbf{H}_0\mathbf{H}_i^T$. Then, $\mathbf{W}$ is computed 
as
$ \mathbf{W} = \mathbf{U}\mathbf{V}^T$.

%The parameters of the mapping function $h(Z_i)$ are obtained by minimizing the 
%above loss function using SGD. 

While we focus on mapping the anchors, the learned mapping function 
is applicable to the whole embedding space.
%, as the vertex embeddings obtained 
%for one subgraph are in the same space. 
In other words, let $\mathbf{W}_i$ be 
the mapping matrix from $\mathbf{H}_i$ to $\mathbf{H}_0$. Then, the reconciled 
embedding space of $\mathbf{F}_i$ is $\mathbf{F}_i\mathbf{W}_i$. 

%Given the set 
%of reconciled spaces $\mathbf{F}_1\mathbf{W}_1, \ldots, 
%\mathbf{F}_n\mathbf{W}_n, \mathbf{F}_0$, we concatenate the matrices and 
%reorder rows (i.e., vertex embeddings), so that the row order (i.e., vertex 
%order) is similar to the one of the reference embedding $\mathbf{E}$ to obtain 
%the reconciled embedding $\mathbf{F}$.

\section{Evaluating Embeddings}
\label{sec:pip}

Having introduced our framework for parallel graph embedding, we turn to the
evaluation of embeddings. Specifically, we assess if and how much an
embedding $\mathbf{E}$ derived by centralised computation differs from an
embedding $\mathbf{F}$ derived by parallel computation. Note that for such a
comparison, embedding $\mathbf{F}$ is
constructed from the reconciled embeddings obtained for the subgraphs, as
explained above: Given the reconciled spaces
$\mathbf{F}_1\mathbf{W}_1, \ldots,
\mathbf{F}_n\mathbf{W}_n, \mathbf{F}_0$, we concatenate the
matrices and reorder rows (vertex embeddings), so that the row order
(vertex order) is similar to the one in $\mathbf{E}$.

%which is created from a parallel embedding
%process, a natural question arises: is $\mathbf{F}$ the same as $\mathbf{E}$?
%If not, how much do they differ?
%Given the reference embedding $\mathbf{Z}_n$ and the parallel embedding
%$\mathbf{Z}_p$, we need a way to compare the distance between them.
%In the following, we aim to answer these questions by discussing a traditional
%approach to compare embeddings and then discuss our proposed approach to
%compare two embeddings leveraging a new metric.

%\subsection{Evaluate embeddings based on downstream tasks}
\subsection{Extrinsic Evaluation}
Extrinsic evaluation measures the quality of embeddings based on some downstream inference tasks.
For instance, in vertex
classification, an embedding is used to assign class labels to vertices after a
learning process. Hence, metrics for classification quality, such as accuracy,
can be used to indirectly quantify the difference between two embeddings.
%Similarly, in the link prediction task which
%involves predicting whether there is an edge between two nodes, metrics such
%as
%prediction accuracy are used to measure the link prediction quality \emph{and}
%the embedding quality.
Such an approach introduces a bias, though, since the inference task may be
influenced by various factors. For instance, in vertex classification, the
quality of the classifier and the correctness of the labels may have a dramatic
impact. Even if the embedding quality is high (it captures the
structural relations of vertices), the classification accuracy may be low, if
the vertex labels are incorrect.
%As a result, measuring the embedding quality using downstream tasks may not
%reflect the embedding quality correctly, hence, it may not truly measure the
%difference between the reference embedding $\mathbf{E}$ and the parallel
%embedding $\mathbf{F}$.

%\subsection{Evaluate embeddings using PIP metric}
\subsection{Intrinsic Evaluation}
\label{sec:eval}
We therefore propose to compare embeddings solely based on their
intrinsic properties. Given two embedding matrices
$\mathbf{Z}_1$ and $\mathbf{Z}_2$ of the same set of vertices $V$, as the
matrices represent vertex embeddings, the difference between
$\mathbf{Z}_1$ and $\mathbf{Z}_2$ is traced back to the difference between
embeddings of a vertex~$u$:
$\mathbf{z}_{1,u} \in \mathbf{Z}_1$ and $\mathbf{z}_{2,u} \in \mathbf{Z}_2$.
Note though, that the absolute value of $u$'s embeddings has no meaning: If
$\mathbf{z}_{1,u}$ and $\mathbf{z}_{2,u}$ differ, they are simply
different representations of the same entity.
However, we can exploit that a graph embedding aims to preserve the
vertex similarity in the graph. Hence, if vertex $v$ is
similar to $u$ in the graph, the embeddings of $v$ and $u$
shall be close. In other words, the distance between $\mathbf{z}_{1,v}$ and
$\mathbf{z}_{1,u}$ and the distance between $\mathbf{z}_{2,v}$ and
$\mathbf{z}_{2,u}$ should be similar: Both $\mathbf{z}_{1,u}$ and
$\mathbf{z}_{2,u}$ are embeddings of $u$, while  $\mathbf{z}_{1,v}$ and
$\mathbf{z}_{2,v}$ are embeddings of $v$.

%In other words, even if $\mathbf{z}_{1,u}$ and $\mathbf{z}_{2,u}$ are different, they should have the same usage.
%This means $\mathbf{z}_{1,u}$ and $\mathbf{z}_{2,u}$ need to be similar in their \emph{applicability}. The applicability of the node embeddings come from the homophily principle of network embedding where nodes that are close in the graph need to have close embeddings in the embedding space.
The above means that the similarity between $\mathbf{z}_{1,u}$ and
$\mathbf{z}_{2,u}$ may be revealed in comparison to other vertices' embeddings.
Let $L= (c_1,\ldots,c_l)$ be a set of `core' vertices. Then, the difference
between $\mathbf{z}_{1,u}$ and $\mathbf{z}_{2,u}$ is measured based on their
distance with the embeddings of the core vertices:
\begin{multline}
d(\mathbf{z}_{1,u}, \mathbf{z}_{2,u} | L) = ||(\mathbf{z}_{1,u} \mathbf{z}^T_{1,c_1},\ldots,\mathbf{z}_{1,u} \mathbf{z}^T_{1,c_l}) - \\ (\mathbf{z}_{2,u} \mathbf{z}^T_{2,c_1},\ldots,\mathbf{z}_{2,u} \mathbf{z}^T_{2,c_l})||_F 	
\end{multline}
where $\mathbf{z}_{1,u} \mathbf{z}^T_{1,c_1}$ is the inner product. Based
thereon, we can compute the difference between two embedding spaces $\mathbf{Z}_1$
and $\mathbf{Z}_2$ based on the difference in embeddings of each
vertex $u \in V$. Considering all vertices as `core'
vertices ($L = V$), we obtain the PIP metric as defined for word embedding
tasks~\cite{yin2018global,yin2018dimensionality}:
%The PIP metric can captured the above observations as it considered all the nodes as core nodes\cite{yin2018global,yin2018dimensionality}, which makes the above equation become:
\[ PIP(\mathbf{Z}_1,\mathbf{Z}_2) = ||\mathbf{Z}_1\mathbf{Z}_1^T -
\mathbf{Z}_2\mathbf{Z}_2^T||_F. \]
The smaller the obtained value, the more similar are the two embedding spaces.
\section{Theoretical Bounds\label{sec:theory}}
Next, we analyse the difference between an embedding obtained with centralised
computation, $\mathbf{E}$, and the one obtained by parallel
computation, $\mathbf{F}$, using the above metric. We first discuss how graph
embedding can be considered as a form of matrix factorisation, before providing
a bound for the aforementioned difference.

\subsection{Graph Embedding as Matrix Factorisation}

Most graph embedding techniques are implicit matrix
factorisations~\cite{qiu2018network,liu2019general} of a \emph{signal matrix}
$\mathbf{M}$ that can be constructed from the adjacency matrix $\mathbf{A}$.
Graph embedding techniques differ in how they construct the signal
matrix~\cite{qiu2018network}.
For instance, the LINE embedding technique~\cite{tang2015line} aims to
factorise the matrix $\mathbf{M} =
({1}/{vol(G)}) \mathbf{C}^{-1}\mathbf{A}\mathbf{C}^{-1}$,where $\mathbf{C}$ is
the degree diagonal matrix and $vol(G)$ is the sum of all vertex degrees. 
Another example is the
signal matrix of HOPE~\cite{ou2016asymmetric}, which is $\mathbf{M} =
\mathbf{A}\mathbf{A}$.

Given a signal matrix $\mathbf{M}$ and its SVD $\mathbf{M} =
\mathbf{U}\mathbf{D}\mathbf{V}$, an embedding is constructed by $\mathbf{E} =
\mathbf{U}_{.,1:k}\mathbf{D}_{1:k,1:k}^{\alpha}$ with $\alpha \in [0,1]$ as a
\emph{signal parameter} and $k$ as the embedding dimensionality.

\subsection{Bound for Parallel Computation}

For simplicity, we assume that the graph is decomposed into two non-overlapping
subgraphs of size $n_1$ and $n_2$. The assumption of a non-overlapping split
does not affect our analysis, since the PIP metric is agnostic to the
application of an orthogonal mapping matrix
$\mathbf{E}\mathbf{E}^T=(\mathbf{EW})(\mathbf{EW})^T$.
%we assume that we only have 2 machines i.e. $n = n_1 + n_2$ where $n$ is the total number of nodes and $n_i$ is the node constraint for the $i$-th machine.

\begin{mythm}
\label{thm:diff}
Let $\mathbf{E}$ and $\mathbf{F}$ be embeddings derived by centralised and
parallel computation, respectively. The PIP distance between
$\mathbf{E}$ and $\mathbf{F}$ is bounded:
%\[||\mathbf{Z}_n\mathbf{Z}_n^T - \mathbf{Z}_p\mathbf{Z}_p^T|| \leq \sum_{m \in \{n_1, n_2\}}\sqrt{\sum_{i=1}^{2m - n}(\lambda_i^{2\alpha} - \lambda_{i+2(n-m)}^{2\alpha})^2 +\sum_{i=2m-\todo{n+1}}^{\todo{l}}\lambda_i^{2\alpha}} + 2\sum_{i=1}^{\todo{l}}\lambda_i^{2\alpha} + i\sqrt{2}\]

%\scriptsize
\vspace{-1.2em}
\begin{multline}
PIP(\mathbf{E},\mathbf{F})\leq 3\sum_{i=1}^{k}\lambda_i^{2\alpha} + 2\sum_{i=1}^k(\lambda_i^{2\alpha} - \lambda_{i+1}^{2\alpha})\sqrt{i(n-i)}	 + \\ + \sum_{m \in \{n_1, n_2\}}\sqrt{\sum_{i=1}^{2m - n}(\lambda_i^{2\alpha} - \lambda_{i+2(n-m)}^{2\alpha})^2 +\sum_{i=2m-n+1}^{k}\lambda_i^{4\alpha}}
\end{multline}

\vspace{-.3em}
\normalsize
\noindent
where $\lambda_i$ is the $i$-th singular value of the signal matrix
$\mathbf{M}$ and $\alpha$ is the signal parameter.
\end{mythm}

\noindent
To prove Theorem~\ref{thm:diff}, we need the following lemmas.
\begin{mylmm}
\label{lmm:sub1}
Let $\lambda_i$ be the $i$-th singular value of a signal matrix $\mathbf{M}$
and $\lambda'_i$ be the $i$-th singular value of a signal matrix $\mathbf{M}'$,
where $\mathbf{M}'$ is the submatrix of $\mathbf{M}$. Then, it holds that:

\vspace{-.5em}
%\scriptsize
\begin{multline*}
\sqrt{\sum_{i=1}^k(\lambda_i^{2\alpha} - \lambda_i'^{2\alpha}) } \leq
\sqrt{\sum_{i=1}^{2m - n}(\lambda_i^{2\alpha} - \lambda_{i+2(n-m)}^{2\alpha})^2 
+ \Theta}	
\end{multline*}

%\normalsize
\noindent
where $\Theta = \sum_{i=2m-n+1}^{k}\lambda_i^{4\alpha}$.
\end{mylmm}

\begin{mylmm}
\label{lmm:sub2}
With $\lambda_i$, $\lambda'_i$,  $\mathbf{M}$,  and $\mathbf{M}'$ as defined in
Lemma~\ref{lmm:sub1}, it holds that:

\vspace{-.5em}
%\scriptsize
\[\sqrt{\sum_{i}^k\lambda_i'^{2\alpha}} \leq \sum_{i=1}^{k}\lambda_i^{2\alpha}\]

\normalsize
\end{mylmm}

\noindent
Both lemmas follow directly from the properties of singular values of
submatrices, see~\cite{thompson1972principal}.
%are the direct result from the Theorem regarding the singular values of
%submatrices~\cite{thompson1972principal}:
%\begin{mythm}
%	Let $\mathbf{M}'$ be the submatrices of matrix $\mathbf{M}$, the singular
%values of $\mathbf{M}'$ is bounded by $\lambda_i \geq \lambda'_i \geq
%\lambda_{i+2(n-n_1)}, \forall i \leq 2n_1 - n$ and $\lambda_i \geq \lambda'_i
%\geq 0, \forall i > 2n_1 -n$ where $n, n_1$ be the size of matrices
%$\mathbf{M}$ and $\mathbf{M}'$ respectively.
%\end{mythm}
Using these results, we prove Theorem~\ref{thm:diff} as follows:
%Note that in the following proof, unless stated otherwise, all the norms are Frobenius norm.

\begin{proof}
	Let $\mathbf{F}_1, \mathbf{F}_2$ be the embeddings of the two subgraphs from
	which $\mathbf{F}$ has been derived. Let further $\mathbf{E}_1,
	\mathbf{E}_2$ be the corresponding partial embeddings from $\mathbf{E}$. 
	This
	means that $\mathbf{E}$ is the embedding derived by decomposing the signal
	matrix $\mathbf{M}$, while $\mathbf{M}_1$ and $\mathbf{M}_2$ are the signal
	matrices of the two subgraphs that generate $\mathbf{F}_1, \mathbf{F}_2$.
	In other words, $\mathbf{F}_1 = \mathbf{Y}\mathbf{D}_1^\alpha, \mathbf{F}_2 = \mathbf{Z}\mathbf{D}_2^\alpha$ and $\mathbf{E}_i = \mathbf{X}_i\mathbf{D}^\alpha$. In addition, $\mathbf{M}_1, \mathbf{M}_2$ are the submatrices of $\mathbf{M}$. We also have:
	\begin{multline}
	\label{equ:pips}
		%||\mathbf{E}_1\mathbf{E}_1^T - \mathbf{F}_1\mathbf{F}_1^T|| +||\mathbf{E}_2\mathbf{E}_2^T - \mathbf{F}_2\mathbf{F}_2^T||	
		PIP(\mathbf{E}, \mathbf{F}) = PIP(\mathbf{E}_1,\mathbf{F}_1) + PIP(\mathbf{E}_2,\mathbf{F}_2)
	\\+ ||\mathbf{E}_1\mathbf{E}_2^T - \mathbf{F}_1\mathbf{F}_2^T|| + ||\mathbf{E}_2\mathbf{E}_1^T - \mathbf{F}_2\mathbf{F}_1^T||
	\end{multline}
%	From Theorem~2~\cite{yin2018dimensionality} and Lemma~\ref{}, we have:
We bound the terms on the right-hand side of the above equation using Theorem~2
in~\cite{yin2018dimensionality}:

\vspace{-1em}
\small
	\begin{multline}
	\label{equ:pip1}
		PIP(\mathbf{E}_1,\mathbf{F}_1) + PIP(\mathbf{E}_2,\mathbf{F}_2) \leq \sum_{j \in \{1,2\}}\sqrt{\sum_{i=1}^k(\lambda_i^{2\alpha} - \lambda_i^{(j)2\alpha})}  \\ + \sqrt{2}\sum_{i=1}^k(\lambda_i^{2\alpha} - \lambda_{i+1}^{2\alpha})(||\mathbf{Y}_{.,1:i}^T\mathbf{X}_{.,i:n}^{(1)}|| + ||\mathbf{Z}_{.,1:i}^T\mathbf{X}_{.,i:n}^{(2)}||)
	\end{multline}

\vspace{-.3em}
\normalsize
\noindent
The first term can be further bounded using Lemma~\ref{lmm:sub1}. The
term $||\mathbf{Y}_{.,1:i}^T\mathbf{X}_{.,i:n}^{(1)}|| +
||\mathbf{Z}_{.,1:i}^T\mathbf{X}_{.,i:n}^{(2)}||$ is bounded as follows:

\vspace{-.5em}
\small
	\begin{multline}
		||\mathbf{Y}_{.,1:i}^T\mathbf{X}_{.,i:n}^{(1)}|| + ||\mathbf{Z}_{.,1:i}^T\mathbf{X}_{.,i:n}^{(2)}|| \leq
		\\ ||\mathbf{Y}_{.,1:i}^T||||\mathbf{X}_{.,i:n}^{(1)}|| + ||\mathbf{Z}_{.,1:i}^T||||\mathbf{X}_{.,i:n}^{(2)}|| \\ \leq \sqrt{i}(||\mathbf{X}_{.,i:n}^{(1)}|| + ||\mathbf{X}_{.,i:n}^{(2)}||) \leq  \sqrt{i}\sqrt{2(||\mathbf{X}_{.,i:n}^{(1)}||^2 + ||\mathbf{X}_{.,i:n}^{(2)}||^2)} \\ \leq \sqrt{i}\sqrt{2(n-i)} \leq \sqrt{2i(n-i)}
	\end{multline}

\vspace{-.0em}
\normalsize
\noindent
The next to last term in Eq.~\ref{equ:pips} is bounded as follows:

\vspace{-1em}
\small
	\begin{multline}
		\label{equ:pip2}
		||\mathbf{E}_1\mathbf{E}_2^T - \mathbf{F}_1\mathbf{F}_2^T|| \leq
		||\mathbf{E}_1\mathbf{E}_2^T|| + ||\mathbf{F}_1\mathbf{F}_2^T||  \\
		\leq ||\mathbf{E}_1||||\mathbf{E}_2^T|| + ||\mathbf{F}_1||||\mathbf{F}_2^T|| \\ \leq \frac{1}{2	}(||\mathbf{E}_1||^2 + ||\mathbf{E}_2^T||^2) + ||\mathbf{F}_1||||\mathbf{F}_2^T|| 
		\\ \leq \frac{1}{2}\sum_{i}^{k}\lambda_i^{2\alpha} + \sqrt{\sum^{k}_{i=1}\lambda^{(1)2\alpha}_i}\sqrt{\sum^{k}_{i=1}\lambda^{(2)2\alpha}_i}
	\end{multline}

\vspace{-.0em}
\normalsize
\noindent
We obtain a similar bound for the last term in Eq.~\ref{equ:pips}.
Combining the above equations, the last two terms in Equation~\ref{equ:pips}
are bounded as follows:

\vspace{-1em}
\small
	\begin{multline}
		\label{equ:pip2}
		||\mathbf{E}_1\mathbf{E}_2^T - \mathbf{F}_1\mathbf{F}_2^T|| + ||\mathbf{E}_2\mathbf{E}_1^T - \mathbf{F}_2\mathbf{F}_1^T|| 	\leq
		\\ \sum_{i}^{k}\lambda_i^{2\alpha} + 2\sqrt{\sum^{k}_{i=1}\lambda^{(1)2\alpha}_i\sum^{k}_{i=1}\lambda^{(2)2\alpha}_i} \leq \sum_{i}^{k}\lambda_i^{2\alpha} + 2\sum_{i=1}^k\lambda_i^{2\alpha}
	\end{multline}

\vspace{-.3em}
\normalsize
\noindent
where we use Lemma~\ref{lmm:sub2} in the last inequality.
Combining Equation~\ref{equ:pips}, \ref{equ:pip1} and \ref{equ:pip2}, we obtain
Theorem~\ref{thm:diff}.
\end{proof}

Theorem~\ref{thm:diff} shows that the PIP distance between embeddings derived
by centralised and parallel computation is bounded only by the singular
values of the original signal matrix and a chosen embedding size. Hence, while
there are different ways to decompose a graph, the quality of
the resulting embedding is bounded by the graph structure.

\section{Empirical evaluation\label{sec:exp}}

\subsection{Experimental setup}

%\begin{table}[h!]
%\end{table}

\sstitle{Datasets}
We evaluate our approach to parallel computation of graph embeddings using 
several datasets (Table~\ref{tbl:stats}) in different tasks. 
%\emph{Cora} is a graph of papers with edges capturing citation dependencies and vertex labels indicating paper categories. 
For the task of vertex classification, we use 
\emph{BlogCatalog}~\cite{tang2009relational}, a social network of blog authors, 
in which labels 
capture topics, and \emph{Flickr}~\cite{tang2009relational}, a network of users 
on a photo sharing website where labels represent user groups having the same 
interest. 
For the task of link prediction, we consider 
\emph{Arxiv}~\cite{leskovec2007graph}, a collaboration graph of scientist, and 
\emph{Youtube}~\footnote{http://networkrepository.com/soc-youtube.php}, a 
network of Youtube users. 
To evaluate the scalability of our proposed method, we rely on 
\emph{LiveJournal}~\cite{yang2015defining}, an online blogging network of user 
friendships, and \emph{Wikipedia}~\cite{yin2017local}, a dataset containing 
hyperlinks from Wikipedia articles. 
%restricted to pages in the top set of categories. 
In total, the LiveJournal dataset contains nearly 40 million edges and nodes, 
while the Wikipedia dataset contains more than 25 million edges and nodes. All 
datasets are publicly available.
%Descriptive statistics of the datasets are provided in . 

%\emph{Youtube}: is a social network between users of the popular video sharing website. The labels here represent groups of viewers that enjoy common video genres (e.g. anime and wrestling).

%It is worth noting that we have the node labels for all the datasets except the Youtube dataset. However, we are still able to evaluate the embedding quality of the Youtube dataset using the PIP metric. 
%We also compare the performance of our method on several synthetic datasets. The synthetic datasets are only used to evaluate the training time of our method and the quality of the embeddings using PIP loss

\sstitle{Metrics} We measure the \emph{efficiency} of parallel computation 
of graph embeddings and the \emph{embedding 
quality}. For our framework, learning time is 
measured as the maximum embedding time on each compute node and the maximum 
time required to reconcile two embedding spaces. The memory requirement is 
measured in the same vein.
The embedding quality is assessed indirectly, as the F1-score in vertex 
classification, and ROC and AP scores in link prediction. A direct quality 
metric is the PIP distance between the embeddings derived by 
centralised and parallel computation (see Section~\ref{sec:eval}), 
normalized by the number of graph vertices.

%\subsection{Baselines} 

\subsection{Efficiency} 
\begin{figure*}
	\begin{minipage}{.22\linewidth}
		\centering
		\includegraphics[width=1.0\linewidth]{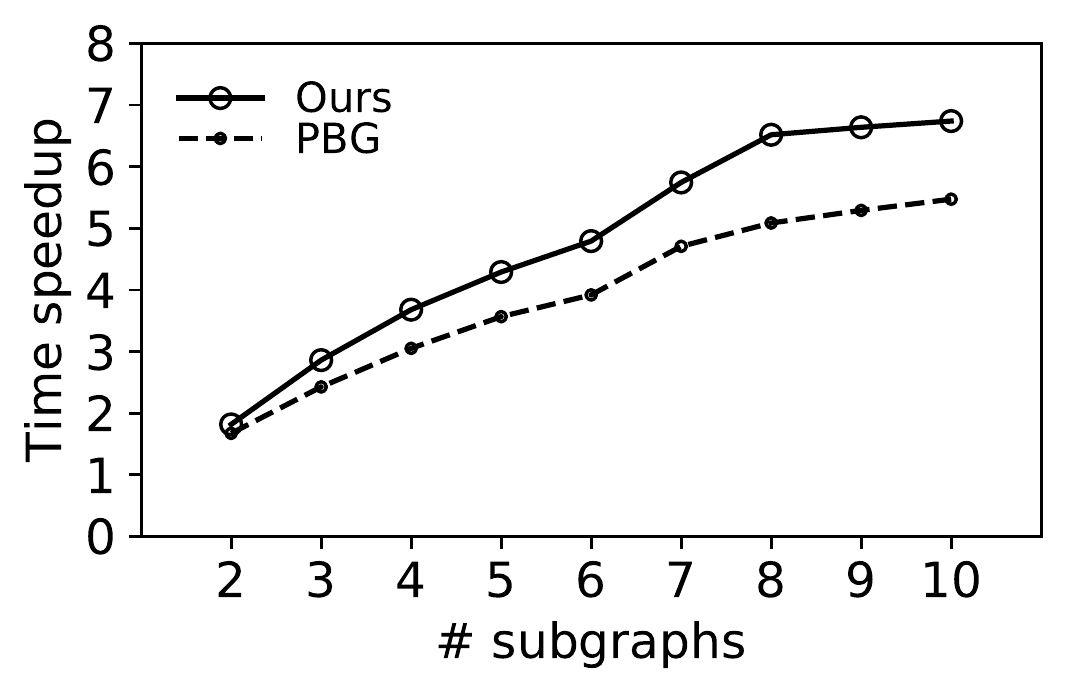}
		\caption{Time speedup}
		\label{fig:time}
	\end{minipage}
	\quad
	\begin{minipage}{.22\linewidth}
		\includegraphics[width=1.0\linewidth]{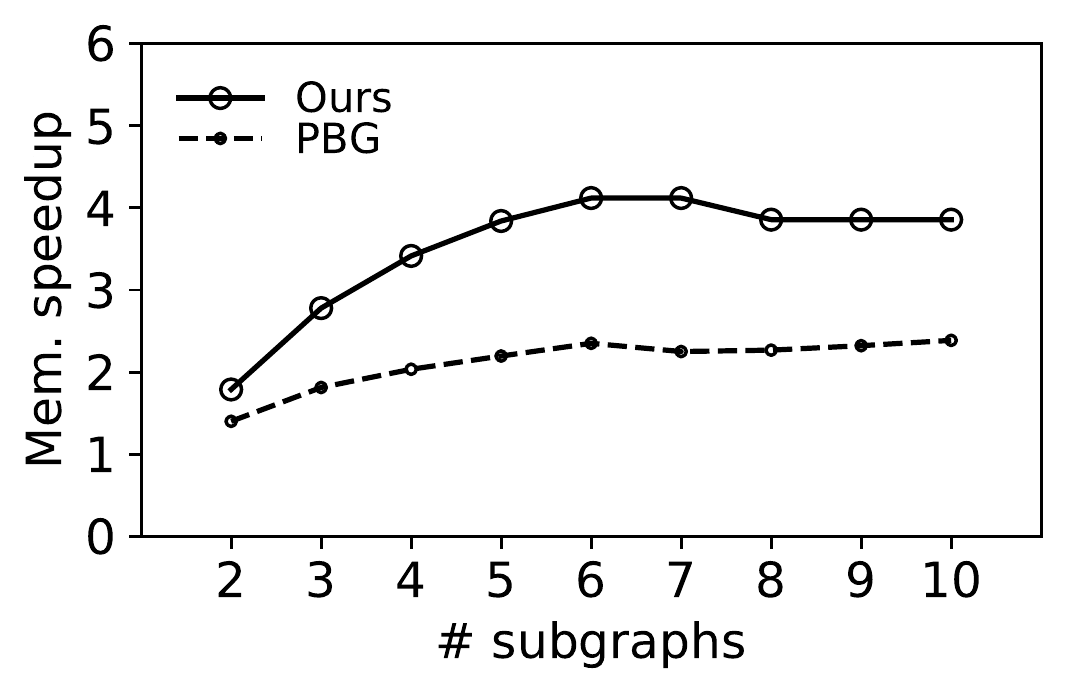}
		\caption{Memory speedup}
		\label{fig:mem}
	\end{minipage}
	\quad
	\begin{minipage}{.49\linewidth}
		\begin{minipage}{.47\linewidth}
			\includegraphics[width=1.0\linewidth]{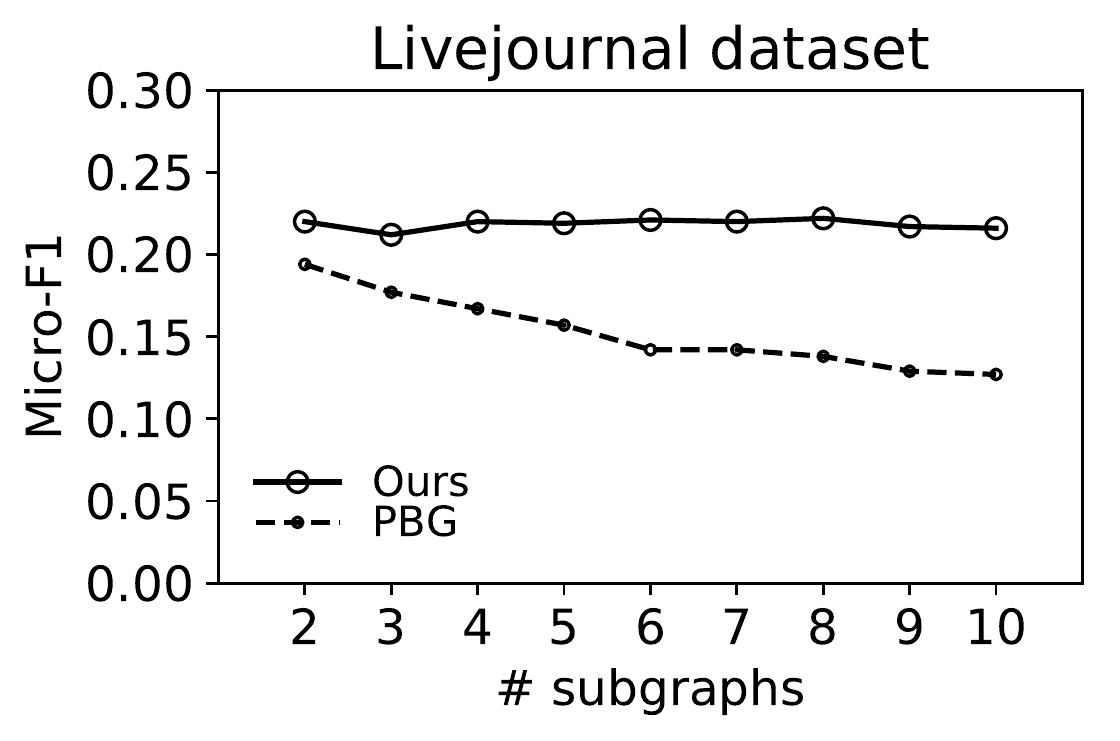}
			 
		\end{minipage}
		\quad
		\begin{minipage}{.47\linewidth}
			\includegraphics[width=1.0\linewidth]{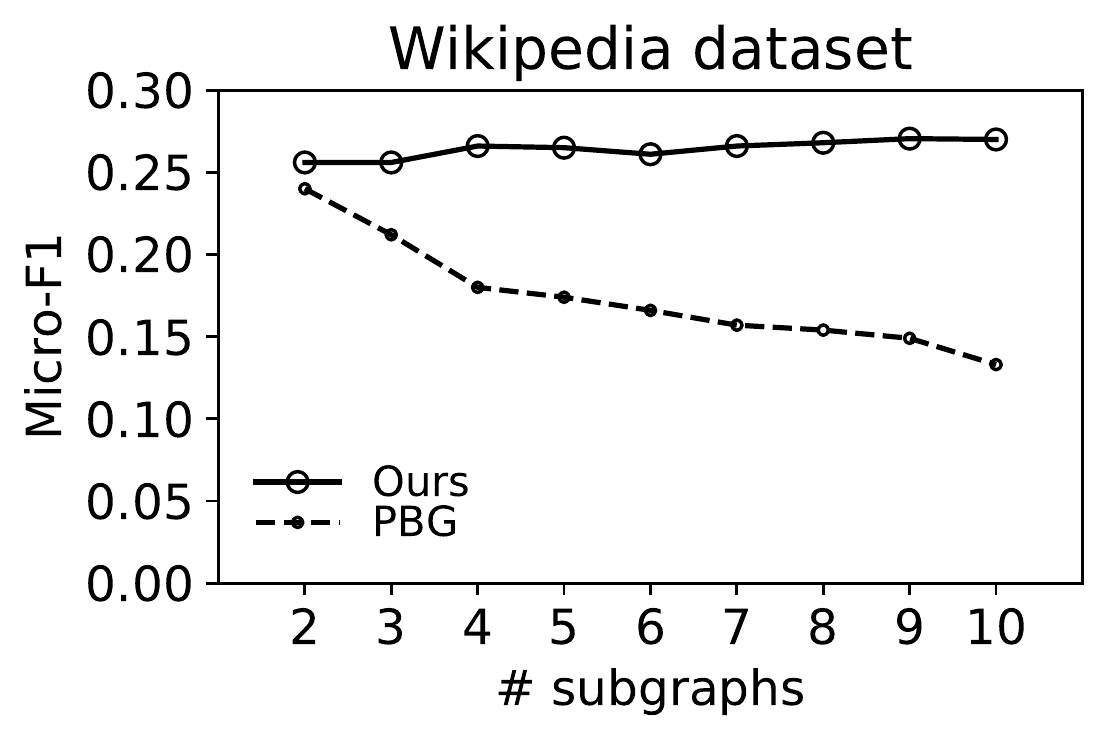}
		\end{minipage}
		\caption{Effects of parallelism on performance.}
		\label{fig:f1}
	\end{minipage}
\end{figure*}

We first explore the gains in efficiency that result from our approach to 
parallel computation. We compare our framework with 
PBG~\cite{lerer2019pytorch}, which is the only existing technique to distribute 
graph embedding to a set of compute nodes. Unlike our 
approach, however, PBG assumes that compute nodes share 
storage and communicate during training. To achieve a fair comparison, we 
disable such communication among nodes when running PBG. Also, note that  
PBG is inherently limited to edge-based embedding 
techniques.
We increase the number 
of compute nodes (and, thus, subgraphs) from 2 to 10 for 
our largest datasets (\emph{Livejournal} and \emph{Wikipedia}), selecting 1\% of the 
vertices as anchors. This 
decreases learning time significantly. For instance, we achieve a 
time speedup of 6.9 with eight compute nodes in comparison with only 5.1 for PBG on the Livejournal dataset (Figure~\ref{fig:time}). 
Our approach outperforms PBG constantly in term of speedup as we increase the number of compute nodes. 
In addition, from Figure~\ref{fig:f1}, we observe that the quality of the 
embeddings remains nearly the same as the level of parallelism increases. On 
the other hand, the embedding quality of PBG decreases significantly as more 
compute nodes are used. For instance, while the difference in micro-f1 score 
between our approach and PBG is only 0.016 when 2 compute nodes are used, it 
becomes 0.137 with 10 compute nodes, which is a nearly 9$\times$ increase. We 
can explain 
this observation based on the reference embedding method. Since PBG is 
restricted to an edge-based embedding method, its embedding quality is more 
susceptible to partitioning. {In this experiment, however, we rely on
DeepWalk as the reference embedding method, which alleviates the above problem: 
Random walks allow to connect subgraphs from their 
common anchors.}

\begin{table}
	%    \vspace{-1em}
	\centering
	%    \scriptsize
	\caption{Datasets statistics\label{tbl:stats}}
	  \vspace{.4em}
	\begin{tabular}{ l l l}
		\toprule
		\textbf{Dataset} & \textbf{\# Vertices} & \textbf{\# Edges} \\
		\midrule
		%Cora            & 2'708           & 5'429                   \\ 	
		BlogCatalog           & 10'312             & 333'983 \\
		Flickr       & 80'513               &  5'899'882  \\
		Arxiv & 18'722 & 198'110 \\
		Youtube & 495'957 & 1'936'748 \\      
		Wikipedia & 1'791'489 & 28'511'807 \\      
		LiveJournal & 3'997'962 & 34'681'189 \\
		\bottomrule
	\end{tabular}
	\vspace{-1.4em}
\end{table}

\begin{table*}[t]
\begin{minipage}{.47\linewidth}
  \caption{Our framework against different embedding techniques (micro-F1). 
  Similar trends hold for macro-F1.}
    %of our framework on different datasets and against
\label{tbl:f1}
\vspace{.4em}
\resizebox{0.9\linewidth}{!}{
\begin{tabular}{@{}lllllll@{}}
\toprule
\multicolumn{1}{c}{\textbf{}} & \multicolumn{3}{c}{\textbf{BlogCatalog}} & \multicolumn{3}{c}{\textbf{Flickr}} \\ 
\cmidrule(l){2-4}                              \cmidrule(l){5-7}
                              & DW           & HOPE        & SGN         & DW         & HOPE       & SGN       \\ 
                              \cmidrule(l){2-7}
NoRecon                             & 0.272        & 0.238       & 0.231       & 0.365      & 0.271      & 0.282     \\ %\midrule
Recon-R                           & 0.272        & \textbf{0.245}       & 0.238       & 0.367      & 0.290      & 0.281     \\
Recon-S                           & \textbf{0.284}        & 0.242       & \textbf{0.246}       & \textbf{0.384}      & \textbf{0.298}      & \textbf{0.29}      \\ \bottomrule
\end{tabular}
}
\end{minipage}
\quad
\begin{minipage}{.47\linewidth}
  \caption{PIP distance to the centralised embeddings (the lower, the better).}
\vspace{.4em}
  \label{tbl:pip}
  \resizebox{0.98\linewidth}{!}{
\begin{tabular}{lllllll}
\hline
\multicolumn{1}{c}{\textbf{}} & \multicolumn{3}{c}{\textbf{BlogCatalog}} & \multicolumn{3}{c}{\textbf{Flickr}} \\ 
\cmidrule(l){2-4}                              \cmidrule(l){5-7}
                              & DW           & HOPE         & SGN        & DW         & HOPE        & SGN      \\                              
                              \cmidrule(l){2-7}
NoRecon                       & 1.55         & 18.6         & 0.42       & 17.06      & 157.42      & 3.12     \\
Recon-R                     & \textbf{1.08}         & 10.57        & 0.38       & 10.31      & 67.11       & \textbf{2.65}     \\
Recon-S                     & 1.102        & \textbf{9.45}         & \textbf{0.35}       & \textbf{9.52}       & \textbf{38.21}       & 2.77     \\ \hline
\end{tabular}
}
\end{minipage}
\end{table*}

\subsection{Ablation Studies}

%\sstitle{Ablation study} 
Next, we compare the quality of embeddings generated by our framework and its 
variants, i.e., with and without reconciliation; and using a random or our 
proposed anchor selection strategy. We denote our framework with the proposed 
anchor selection as Recon-S, with random anchor selection as 
Recon-R. We also consider a variant where no reconciliation is used (NoRecon). 
In addition, we analyse the effects of the embedding method on the 
embedding quality. Here, we consider three embedding methods 
(DeepWalk~\cite{perozzi2014deepwalk}, HOPE~\cite{ou2016asymmetric}, and 
SGN~\cite{wu2019simplifying}) that represent the state-of-the-art for the three 
main classes of approaches for graph embedding (walk-based, matrix 
factorization, and graph neural network).

\begin{table*}[!ht]
\centering
\caption{Effects of embedding strategies on link prediction.}
\label{tbl:link}
\vspace{.4em}
\begin{tabular}{cllllllllllll}
\hline
\multirow{3}{*}{\textbf{}} & \multicolumn{6}{c}{\textbf{arxiv}}                                                                         & \multicolumn{6}{c}{\textbf{youtube}}                                                                 \\                               \cmidrule(l){2-7} \cmidrule(l){8-13}
                           & \multicolumn{2}{c}{DW} & \multicolumn{2}{c}{HOPE}       & \multicolumn{2}{c}{SGN}                          & \multicolumn{2}{c}{DW} & \multicolumn{2}{c}{HOPE} & \multicolumn{2}{c}{SGN}                          \\
                           \cmidrule(l){2-3} \cmidrule(l){4-5} \cmidrule(l){6-7} \cmidrule(l){8-9} \cmidrule(l){10-11} \cmidrule(l){12-13}
                           & ROC        & AP        & ROC   & \multicolumn{1}{c}{AP} & \multicolumn{1}{c}{ROC} & \multicolumn{1}{c}{AP} & ROC        & AP        & ROC         & AP         & \multicolumn{1}{c}{ROC} & \multicolumn{1}{c}{AP} \\
                           \cmidrule{2-13}
D                          & 0.843      & 0.883     & 0.890 & 0.926                  & 0.827                   & 0.877                  & 0.822      & \textbf{0.845}     & 0.813       & 0.880      & 0.622                   & 0.699                  \\ 
DAB                        & 0.808      & 0.856     & 0.887 & 0.924                  & 0.815                   & 0.898                  & 0.749      & 0.776     & 0.810       & 0.876      & 0.641                   & 0.706                  \\
DAS                        & \textbf{0.878}      & \textbf{0.899}     & \textbf{0.918} & \textbf{0.94}                   & \textbf{0.857}                   & \textbf{0.901}                  & \textbf{0.831}      & 0.842     & \textbf{0.843}       & \textbf{0.896}      & \textbf{0.662}                   & \textbf{0.729}                  \\ \hline
\end{tabular}
\end{table*}

\sstitle{Vertex Classification}
We report the classification accuracy 
(Table~\ref{tbl:f1}) and 
the PIP metric (Table~\ref{tbl:pip}). 
%As expected, embeddings generated by centralised computation yield the best results. 
%In Table~\ref{tbl:f1}, the respective micro and macro-F1 scores 
%are generally the highest.
%the highest across different datasets and embedding techniques. 
%However, the difference regarding the results obtained with our framework are very small, i.e., less than 0.07. 
% for the \emph{BlogCatalog} dataset with the DeepWalk embedding technique.
We observe that our proposed reconciliation mechanism turns out to be 
effective, improving the embedding quality in nearly all settings. 
For instance, the F1 score for the \emph{Flickr} dataset using the DeepWalk 
embedding 
technique without reconciliation is 0.365, while it is 0.384 when 
reconciliation is used. 
The effect of reconciliation is also seen in Table~\ref{tbl:pip}: The 
differences in PIP distance between the embeddings with centralised computation are smaller for the reconciled embeddings (Recon-R, Recon-S) than for the 
non-reconciled ones (NoRecon). 
Moreover, our anchor selection strategy (Recon-S) provides better 
embeddings 
across 
different datasets and techniques in comparison with the baseline strategy 
(Recon-S). 
% Although there are cases where the F1-scores of our strategy are lower 
% than those of the baseline, the PIP distance between embeddings generated by 
% our strategy and the centralised embeddings are always lower than those 
% observed in relation to the baseline embeddings. For instance, the 
% micro-F1 score of the non-reconciled embedding for the \emph{Cora} dataset 
% using the Deepwalk embedding is 
% 0.801, which is higher than 0.779 of the reconciled embedding, its PIP distance 
% is higher than that of the reconciled embedding. 

\sstitle{Link prediction} 
To evaluate our approach in the context of link prediction, we remove 50\% of 
the edges from the graphs, and use the embeddings to 
predict the removed edges. We select the same amount of vertex pairs that 
have no link in the original graphs as negative samples. The obtained results, 
see Table~\ref{tbl:link}, confirm the observations obtained for 
vertex classification. Our framework consistently outperforms the baseline 
strategies. 
For instance, reconciliation improves the ROC and AP scores (dataset 
\emph{Arxiv}, embedding technique \emph{DeepWalk}) from 0.843 and 
0.883 to 0.878 and 0.899, respectively.

\section{Related Work}
\label{sec:related}
Graph representation learning aims to construct a
low-dimensional model of the vertices in the graph that incorporates the graph 
structure~\cite{cai2018comprehensive,hamilton2017representation}. Existing 
techniques differ in how they map a vertex into an embedding space, and in the 
structural properties that shall be retained. 
There are three main approaches to embed a vertex, using shallow or deep 
encoders~\cite{hamilton2017representation}, or matrix factorisation. Techniques that use shallow 
encoders~\cite{perozzi2014deepwalk,grover2016node2vec} consider vertices as words and random walks as sentences, which allows them to use neural word embedding techniques such as word2vec~\cite{mikolov2013efficient} to construct word/vertex embeddings.
On the other hand, deep encoder approaches such as GraphSAGE~\cite{hamilton2017inductive} or SGN~\cite{wu2019simplifying} consider the neighbourhood of a vertex to generate its embedding. As a consequence, both vertex features and the graph structure may be captured. 
% Most
% approaches~\cite{perozzi2014deepwalk,grover2016node2vec} follow a
% random-walk-based approach to capture a vertex' neighbourhood. Two vertices are
% considered to be close, if one occurs on the random walk from the other.
% %Random-walk-based shallow graph embedding can be considered as an application
% %of word embedding in the graph domain. 
% Other
% approaches~\cite{cao2015grarep,ou2016asymmetric} consider two vertices to be 
% similar, if their neighbourhoods are highly
% overlapping, which is referred to as second-order vertex similarity.
% The focus of existing techniques is on the embedding quality, though. To date,  
% none of these techniques can be distributed in a cluster of compute nodes with 
% resource constraints. 

% Our work also relates to frameworks for parallel and distributed 
% learning~\cite{recht2011hogwild,dean2012large}. These techniques, however, 
% realise model parallelism and distribute computations, such as stochastic 
% gradient descent. Our framework is orthogonal to these approaches, as it can be 
% seen as a form of data parallelism. As such, it is possible to further 
% distribute the computation per subgraph as obtained with our work by exploiting 
% the aforementioned frameworks. \todo{Add discussion on pytorch biggraph}
%into several subgraphs where the computation of the embedding of each subgraph 
%is further distributed using the above techniques. 

There are recent works that aim to scale traditional graph embedding methods to very large graphs. MILE~\cite{mile} aims to scale traditional graph embedding techniques to large graph using only one workstation. A large graph is coarsened into a small one and embedding is performed on the small graph. This approach does not try to leverage a set of compute nodes, though. Closest to our work is PBG~\cite{lerer2019pytorch}, which aims to learn embeddings in a {distributed} manner. As mentioned earlier, PBG assumes a different setting compared to our approach. PBG has been proposed for compute nodes with shared storage that also communicate during training, whereas we focus on a shared-nothing infrastructure, as it is commonly encountered in compute clusters. A second major difference is that
PBG is tailored to scale shallow embedding methods that use negative sampling to large graphs. Our framework, in turn, allows to scale \emph{any} embedding technique and is, therefore, more generally applicable. In particular, it can be combined with deep approaches, such as GraphSAGE and SGN, which suffer from long training times in practice. 

% There are several differences between our framework and PBG. First, PBG and our framework aim for a different setting. PBG assumes a setting with shared storage and communication between compute nodes during training, whereas our setup considers compute nodes to be completely independent during training. Second, PBG is tailored to scale shallow embedding methods that use negative sampling to large graphs. Our framework, in turn, allows to scale \emph{any} embedding technique, which also includes deep approaches, such as GraphSAGE and SGN, which take a lot of time to train.

\section{Conclusion\label{sec:con}}
In this paper, we took on the challenge of scaling arbitrary graph embedding techniques to very large graphs, using a cluster of compute nodes. We showed how to decompose a 
graph for distributed computation, taking into account the resource constraints of a given set of compute nodes. 
%That is we identify the problem of graph embedding using different machines 
%while each machine has different resource constraints such as time and memory. 
%To address this problem, we propose to partition the graph into several 
%smaller 
%subgraphs that can be handled on each machine according to its constraints. 
We then proposed a mechanism to reconcile the embeddings obtained independently for the individual subgraphs and also contributed a way to assess 
embedding quality, independent of a specific inference task. Using this measure, we gave formal guarantees on the difference between embeddings derived by centralised and 
parallel computation. Experiments showed that 
our approach is efficient and effective: It scales well, largely 
maintains the embedding quality, and consistently outperforms the sole existing 
approach to exploit a cluster of compute nodes for graph embedding.

%\bibliographystyle{ACM-Reference-Format}
%\bibliographystyle{named}
%\bibliographystyle{abbrv}
%\bibliography{sample-bibliography}

\end{document}